\documentclass[runningheads]{llncs}
\usepackage[T1]{fontenc}
\usepackage{graphicx}
\usepackage{subfig}
\DeclareUnicodeCharacter{2212}{-}
\usepackage{enumerate}
\usepackage{algorithm2e}

\begin{document}
\title{Cooperate or Compete: A New Perspective on Training of Generative Networks }
\titlerunning{CEN : Cooperatively Evolving Networks}

\author{
Ch. Sobhan Babu\inst{1}
\and
Ravindra Guravannavar\inst{2}
\and
Arvind Hulgeri\inst{3}
}
\authorrunning{Sobhan Babu et al.}
%

\institute{Department of Computer Science and Engineering, Indian Institute of Technology, Hyderabad, India \\
\email{ sobhan@cse.iith.ac.in}
\and
Volunteer, Ramakrishna Mission Ashrama and 
Rajalakshmi Children Foundation, Belagavi, India \\
\email{ravindrag@gmail.com}
\and
Independent Consultant, Pune, India \\
\email{arvind.hulgeri@gmail.com}}

\maketitle              
\begin{abstract}
GANs have two competing modules: the generator module is trained to generate new examples, and the discriminator module is trained to discriminate real examples from generated examples. The training procedure of GAN is modeled as a finitely repeated simultaneous game. Each module tries to increase its performance at every repetition of the base game (at every batch of training data) in a non-cooperative manner. We observed that each module can perform better and learn faster if training is modeled as an infinitely repeated simultaneous game. At every repetition of the base game (at every batch of training data) the stronger module (whose performance is increased or remains the same compared to the previous batch of training data) cooperates with the weaker module (whose performance is decreased compared to the previous batch of training data) and only the weaker module is allowed to increase its performance. 

\end{abstract}
\section{Introduction}
Game theory is the mathematical study of how rational agents  (players) select strategies in different strategic situations in the face of competing strategies acted out by other agents. Game theory assumes that players take rational decisions at all times.  A {\it Simultaneous game} is a game where each agent (player) chooses their action (strategy) to maximize his/her payoff competitively.  In these games, both agents select their strategies in a simultaneous manner. A {\it Sequential-move game} is a game where the players take turns while selecting their strategies, as in chess or negotiations. A game is called  {\it One-shot game} if it is played only once. A game is called  {\it Repeated game} if a base game is played more than once; either a finite or infinite number of times.  In {\it Repeated games}  a base game is played over and over again at discrete time periods.  In a repeated game players make decisions in full knowledge of the history of the game played so far (i.e. the actions chosen by each player in each previous time periods).

The repeated game can be infinite or finite. In a finite repetitive game, both players will expect the other player to choose their action (strategy) to maximize his/her payoff  competitively in the very last base game in the series. Knowing this, both players choose their action (strategy) to maximize his/her payoff competitively in the second-to-last base game. But since both players know that will be the optimal strategy, they will choose their strategy to maximize his/her payoff in the base game before that, and so on, until the very first game. This means in every base game they choose their strategy competitively. In an infinite repetitive game, the base game is repeated with no known end, and competition may not be the best strategy: players may get better payoffs in the long run by cooperating.

{\it Generative adversarial networks} (GANs) are the most recent invention in deep learning. GANs can create new data instances (samples) that resemble the training (ground-truth) data set i.e. learn the probability distribution of training data and generate samples from this probability distribution. For example, GANs can create pictures that look highly similar to photographs of human faces.

\section{Related Work}

A {\it Generative Adversarial Network} (GAN) is a generative deep learning model that trains two modules (generator, discriminator).  The generator tries to learn the probability distribution of the ground-truth data set (training data set), and the discriminator tries to estimate the probability of a given sample is from the ground-truth data set rather than generated by the generator.  This approach corresponds to a two-player finite repeated simultaneous game \cite{Goodfellow14}. Nash Equilibrium is a solution concept in game theory that determines the equilibrium solution in non-cooperative games. This concept gives an equilibrium strategy (equilibrium solution) for a non-cooperative game from which each player lacks any incentive to change (assuming the others also don't change) \cite{Osborne1994}. In game theory, a correlated equilibrium is a solution concept that is more general than the  Nash equilibrium. It was first discussed by mathematician Robert Aumann in 1974 \cite{Aumann}. The idea is that each player chooses their action/strategy according to their private observation of the value of the same public signal. If no player would want to deviate from their strategy (assuming the others also don't deviate), the distribution from which the signals are drawn is called a correlated equilibrium. This talks more about the advantages of cooperation rather than competition.

\section{How GANs Work}

Generative modeling is a learning methodology in deep learning that involves learning the probability distribution of the (ground-truth data set) input data set in such a way that the model can be used to generate new samples (samples from the learned input data set probability distribution) that plausibly could have been drawn from the input data set.

GANs are a  way of training a generative model by framing the problem as a supervised learning problem with two sub-modules (generator, discriminator). The {\it generator module} is trained to generate new samples, and the {\it discriminator module} is  trained to classify samples as either real (from the input data set) or fake (generated by the generator module). This training process is modeled as a finite repetitive simultaneous game  \cite{Goodfellow14}.  In the process of training  GANs, for each batch of  input data set (ground-truth data set), we perform training of both generator and discriminator. A detailed description of the training of a GAN is given in the Algorithm \ref{alg:one}. 

\SetKwComment{Comment}{/* }{ */}
\RestyleAlgo{ruled}
\LinesNumbered
\begin{algorithm}
\DontPrintSemicolon
\SetAlgoLined
\SetNoFillComment
\caption{GAN Training}\label{alg:one}
\KwData{ Ground-Truth Data Set}
\KwResult{Trained GAN}

\For{number of training epochs}{
    \For{number of batches}{
    \;
    \tcc{Training Discriminator}
    Let $z$ be a random sample from some probability distribution (mostly uniform distribution) \;
    $G(z) \gets$ $generator's$ output with $z$ as the input\;
    Let $X$ be the current batch of input data set\;
    $Dx \gets$  $discriminator's$ error of misclassifying elements of $X$ as generated by the $generator$\;
    $Dz \gets$ $discriminator's$ error of misclassifying elements of $G(z)$ as elements of $X$  \;
    Update $discriminator's$ parameters to minimize $Dx+Dz$\;
    \;
    \tcc{Training Generator}
     Let $z$ be a random sample from some probability distribution (mostly uniform distribution)\;
    $G(z) \gets$  $generator's$ output with $z$ as the input\;
    $Dz \gets$ $discriminator's$ error of misclassifying elements of $G(z)$ as elements of $X$  \;
    Update $generators's$ parameters to maximize $Dz$\;
    }
}
\end{algorithm}

In this way, the two modules are competing against each other, they are adversarial in the game theory sense, and are playing a finite repetitive simultaneous game.

\section{Cooperatively Evolving Networks}
The training procedure of GAN is modeled as a finitely repeated simultaneous
game. Each module tries to increase its performance at every repetition
of the base game (at every batch of training data) in a non-cooperative
manner. We observed that each module can perform better and learn
faster if training is modeled as an infinitely repeated simultaneous game.
At every repetition of the base game (at every batch of training data) the
stronger module (whose performance is increased or remains the same
compared to the previous batch of training data) cooperates with the
weaker module (whose performance is decreased compared to the previous batch of training data) and only the weaker module is allowed to
increase its performance. We call generative networks training in this manner as cooperatively evolving network ({\it CEN}).

In the process of training of cooperatively evolving network ({\it CEN}), for each batch of  input data set (ground-truth data set), we perform training of generator and/or discriminator as in the Algorithm \ref{alg:two}. In each iteration, the stronger module (whose performance is increased or remains the same compared to the previous batch of training data) cooperates with  the weaker module (whose performance is decreased compared to the previous batch of training data) and  only the weaker module is allowed to increase its performance. In this way, the two modules cooperate with each other (i.e stronger one is helping the weaker to become strong).

\begin{algorithm}
\DontPrintSemicolon
\SetAlgoLined
\SetNoFillComment
\caption{CEN Training}\label{alg:two}
\KwData{ Ground-Truth Data Set}
\KwResult{Trained CEN}

\For{number of training epochs}{
    \For{number of batches}{
    \;
    \tcc{Computing Generator and Discriminator Errors for this Iteration}
    Let $z$ be a random sample from some probability distribution (mostly uniform distribution) \;
    $G(z) \gets$  $generator's$ output with $z$ as the input\;
    Let $X$ be the current batch of input data set\;
    $Dx \gets$  $discriminator's$ error of misclassifying elements of $X$ as generated by the $generator$\;
    $Dz \gets$ $discriminator's$ error of misclassifying elements of $G(z)$ as elements of $X$  \;
    $generator-error \gets Dz$\;
    $discriminator-error \gets Dz+Dx$\;
    
    \;
    \tcc{Training Generator}
    \If{$generator-error$ is decreased compared to the previous iteration}
    {
     Let $z$ be a random sample from some probability distribution (mostly uniform distribution)\;
    $G(z) \gets$  $generator's$ output  with $z$ as the input\;
    $Dz \gets$ $discriminator's$ error of misclassifying elements of $G(z)$ as elements of $X$  \;
    Update $generators's$ parameters to maximize $Dz$\;
    }
    
    \;
    \tcc{Training Discriminator}
    \If{$discriminator-error$ is increased compared to the previous iteration}
    { 
    Let $z$ be a random sample from some probability distribution (mostly uniform distribution) \;
    $G(z) \gets$ $generator's$ output with $z$ as the input\;
    Let $X$ be the current batch of input data set\;
    $Dx \gets$  $discriminator's$ error of misclassifying elements of $X$ as generated by the $generator$\;
    $Dz \gets$ $discriminator's$ error of misclassifying elements of $G(z)$ as elements of $X$  \;
    Update $discriminator's$ parameters to minimize $Dx+Dz$\;
    }
    
    }
}
\end{algorithm}

\section{Results}
\subsection{Sine Wave}
\begin{figure}
\begin{tabular}{cccc}
\subfloat[100 Epochs]{\includegraphics[width = 0.9in]{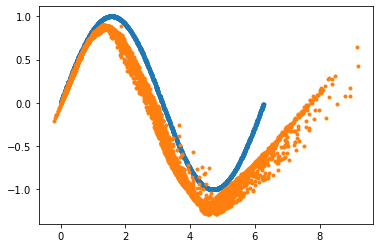}} &
\subfloat[200 Epochs]{\includegraphics[width = 0.9in]{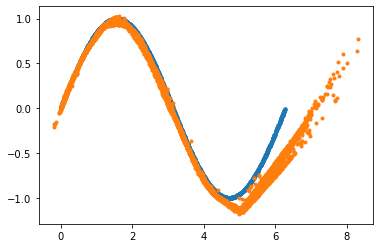}} &
\subfloat[300 Epochs]{\includegraphics[width = 0.9in]{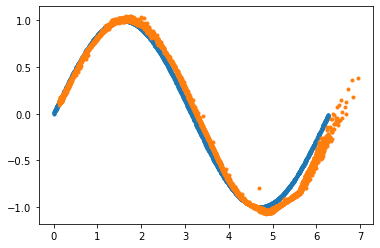}} &
\subfloat[400 Epochs]{\includegraphics[width = 0.9in]{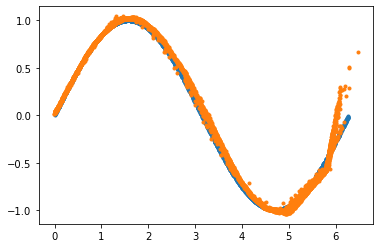}}\\
\subfloat[500 Epochs]{\includegraphics[width = 0.9in]{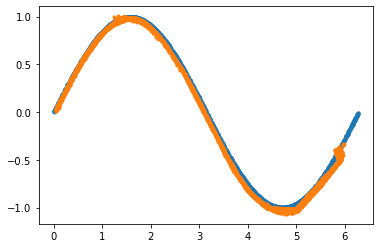}} &
\subfloat[600 Epochs]{\includegraphics[width = 0.9in]{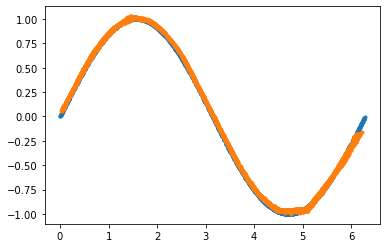}} &
\subfloat[700 Epochs]{\includegraphics[width = 0.9in]{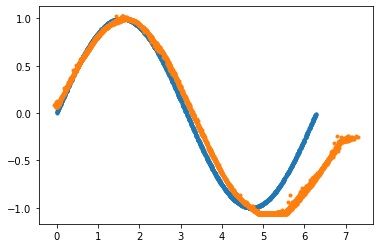}} &
\subfloat[800 Epochs]{\includegraphics[width = 0.9in]{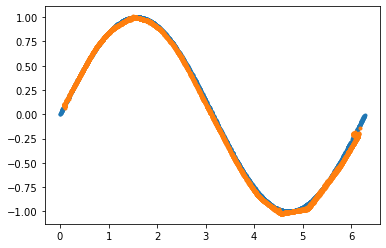}}
\end{tabular}
\caption{Generated by GAN}
\label{GAN}
\end{figure}

\begin{figure}
\begin{tabular}{cccc}
\subfloat[100 Epochs]{\includegraphics[width = 0.9in]{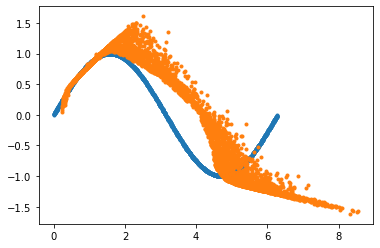}} &
\subfloat[200 Epochs]{\includegraphics[width = 0.9in]{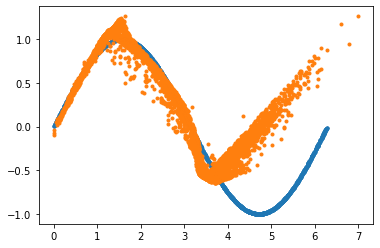}} &
\subfloat[300 Epochs]{\includegraphics[width = 0.9in]{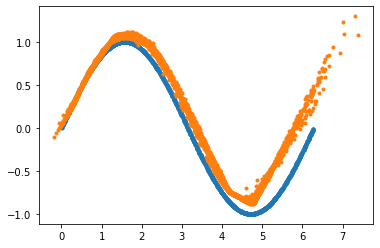}} &
\subfloat[400 Epochs]{\includegraphics[width = 0.9in]{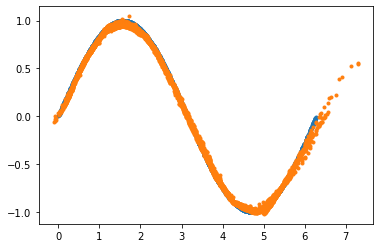}}\\
\subfloat[500 Epochs]{\includegraphics[width = 0.9in]{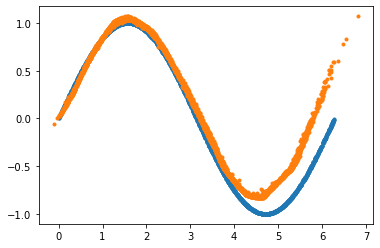}} &
\subfloat[600 Epochs]{\includegraphics[width = 0.9in]{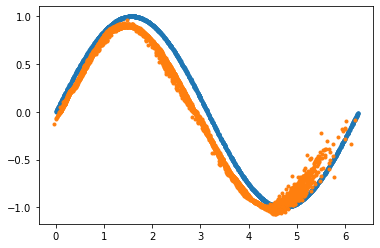}} &
\subfloat[700 Epochs]{\includegraphics[width = 0.9in]{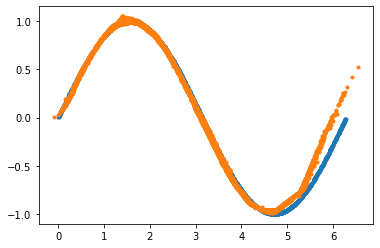}} &
\subfloat[800 Epochs]{\includegraphics[width = 0.9in]{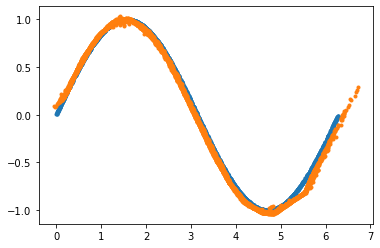}}

\end{tabular}
\caption{Generated by CEN}
\label{CEN}
\end{figure}

The ground-truth data set is a sine wave as shown in blue colour in each image. The orange-colored data set is the sample data generated by the generator. The Figure \ref{GAN} shows the sample data generated for  different numbers of epchos (from one hundred to eight hundred) by GAN. The Figure \ref{CEN} shows the sample data generated for different numbers of epchos (from one hundred to eight hundred) by CEN. 


\subsection{Overlapping Ellipses}
\begin{figure}
\begin{tabular}{cccc}
\subfloat[100 Epochs]{\includegraphics[width = 0.9in]{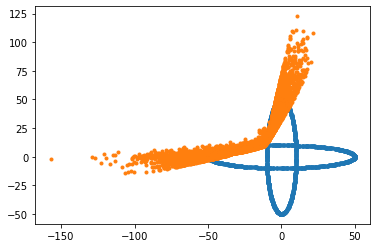}} &
\subfloat[200 Epochs]{\includegraphics[width = 0.9in]{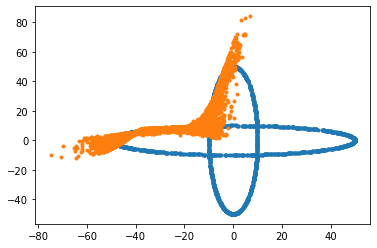}} &
\subfloat[300 Epochs]{\includegraphics[width = 0.9in]{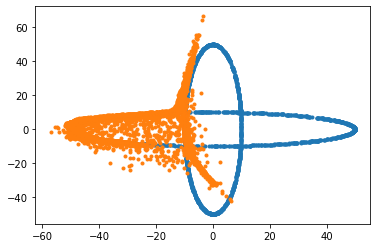}} &
\subfloat[400 Epochs]{\includegraphics[width = 0.9in]{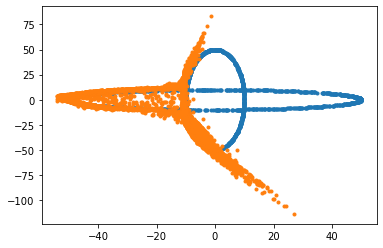}}\\
\subfloat[500 Epochs]{\includegraphics[width = 0.9in]{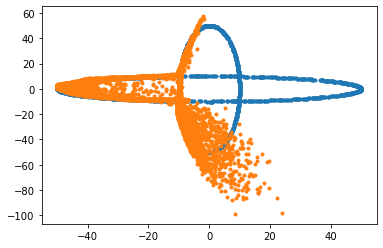}} &
\subfloat[600 Epochs]{\includegraphics[width = 0.9in]{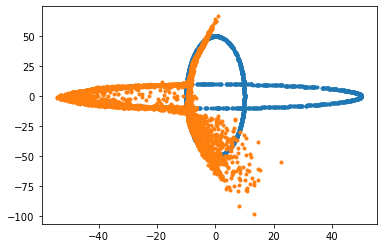}} &
\subfloat[700 Epochs]{\includegraphics[width = 0.9in]{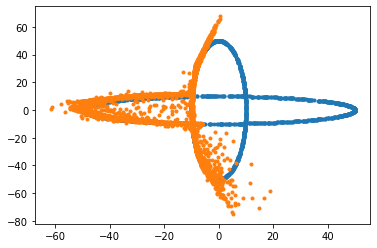}} &
\subfloat[800 Epochs]{\includegraphics[width = 0.9in]{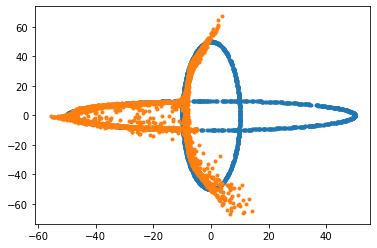}}
\end{tabular}
\caption{Generated by GAN}
\label{eGAN}
\end{figure}

\begin{figure}
\begin{tabular}{cccc}
\subfloat[100 Epochs]{\includegraphics[width = 0.9in]{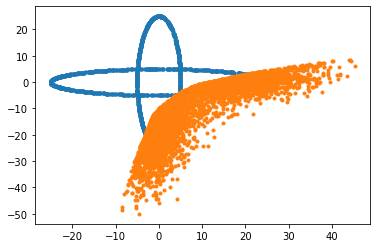}} &
\subfloat[200 Epochs]{\includegraphics[width = 0.9in]{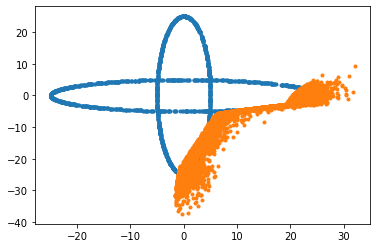}} &
\subfloat[300 Epochs]{\includegraphics[width = 0.9in]{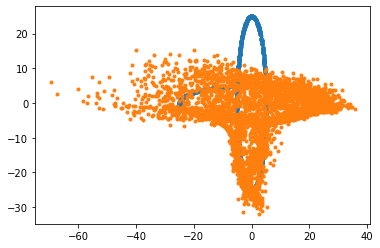}} &
\subfloat[400 Epochs]{\includegraphics[width = 0.9in]{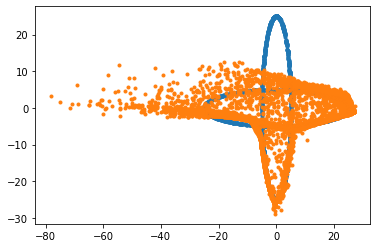}}\\
\subfloat[500 Epochs]{\includegraphics[width = 0.9in]{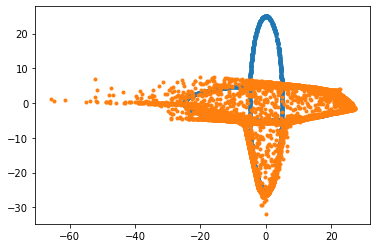}} &
\subfloat[600 Epochs]{\includegraphics[width = 0.9in]{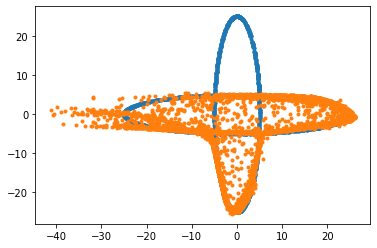}} &
\subfloat[700 Epochs]{\includegraphics[width = 0.9in]{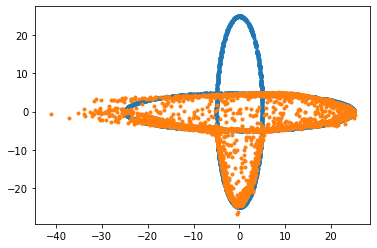}} &
\subfloat[800 Epochs]{\includegraphics[width = 0.9in]{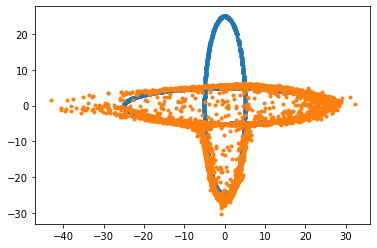}}

\end{tabular}
\caption{Generated by CEN}
\label{eCEN}
\end{figure}

The ground-truth data set is two overlapping ellipses as shown in blue colour in each image. The orange-colored data set is the sample data generated by the generator. The Figure \ref{eGAN} shows the sample data generated for different numbers of epchos (from one hundred to eight hundred) by GAN. The Figure \ref{eCEN} shows the sample data generated for different numbers of epchos (from one hundred to eight hundred) by CEN. 

\subsection{Concentric Circles}
\begin{figure}
\begin{tabular}{cccc}
\subfloat[100 Epochs]{\includegraphics[width = 0.9in]{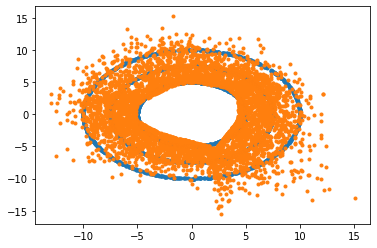}} &
\subfloat[200 Epochs]{\includegraphics[width = 0.9in]{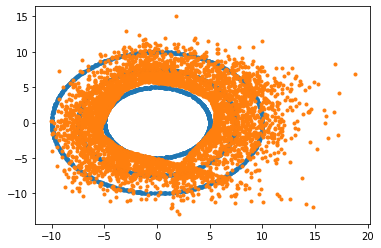}} &
\subfloat[300 Epochs]{\includegraphics[width = 0.9in]{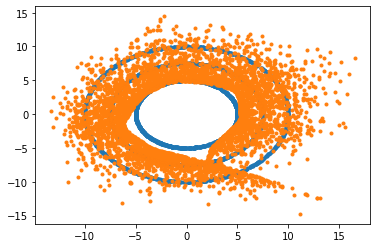}} &
\subfloat[400 Epochs]{\includegraphics[width = 0.9in]{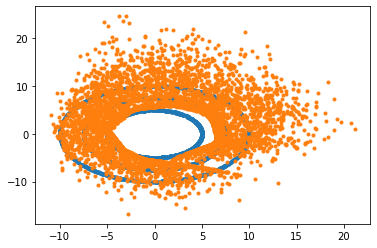}}\\
\subfloat[500 Epochs]{\includegraphics[width = 0.9in]{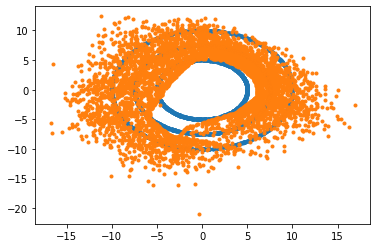}} &
\subfloat[600 Epochs]{\includegraphics[width = 0.9in]{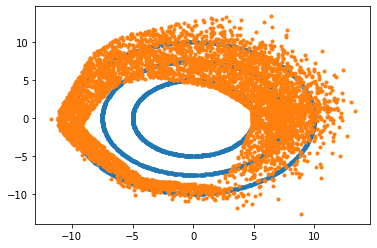}} &
\subfloat[700 Epochs]{\includegraphics[width = 0.9in]{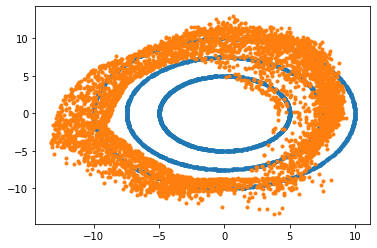}} &
\subfloat[800 Epochs]{\includegraphics[width = 0.9in]{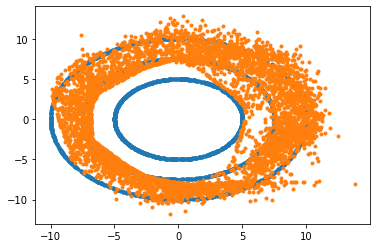}}
\end{tabular}
\caption{Generated by GAN}
\label{cirGAN}
\end{figure}

\begin{figure}
\begin{tabular}{cccc}
\subfloat[100 Epochs]{\includegraphics[width = 0.9in]{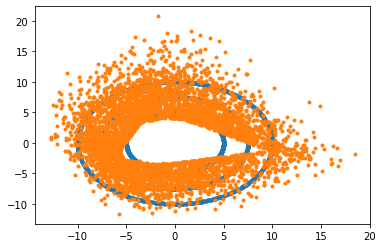}} &
\subfloat[200 Epochs]{\includegraphics[width = 0.9in]{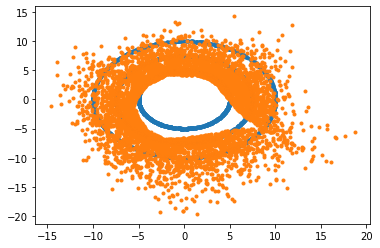}} &
\subfloat[300 Epochs]{\includegraphics[width = 0.9in]{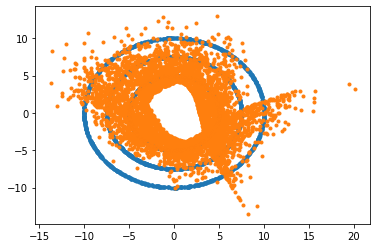}} &
\subfloat[400 Epochs]{\includegraphics[width = 0.9in]{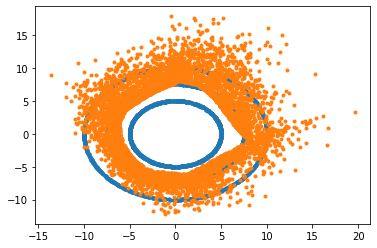}}\\
\subfloat[500 Epochs]{\includegraphics[width = 0.9in]{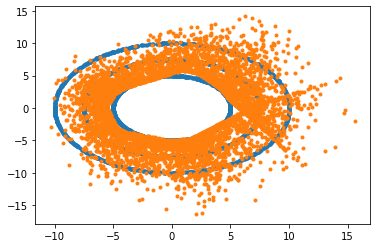}} &
\subfloat[600 Epochs]{\includegraphics[width = 0.9in]{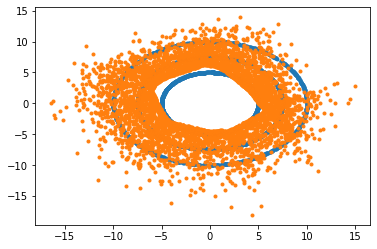}} &
\subfloat[700 Epochs]{\includegraphics[width = 0.9in]{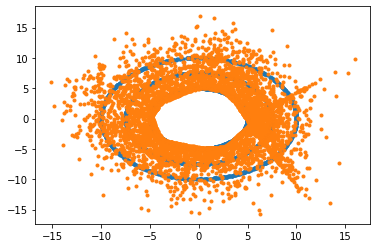}} &
\subfloat[800 Epochs]{\includegraphics[width = 0.9in]{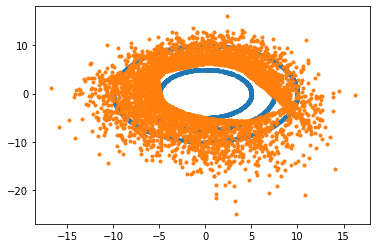}}

\end{tabular}
\caption{Generated by CEN}
\label{cirCEN}
\end{figure}

The ground-truth data set is three concentric as shown in blue colour in each image. The orange-colored data set is the sample data generated by the generator. The Figure \ref{cirGAN} shows the sample data generated for different numbers of epchos (from one hundred to eight hundred) by GAN. The Figure \ref{cirCEN} shows the sample data generated for different number of epchos (from one hundred to eight hundred) by CEN. 

\subsection{Comparison of above three using Jensen–Shannon divergence and computation time}
\begin{figure}
\begin{tabular}{ccc}
\subfloat[Sine Wave]{\includegraphics[width = 1.5in]{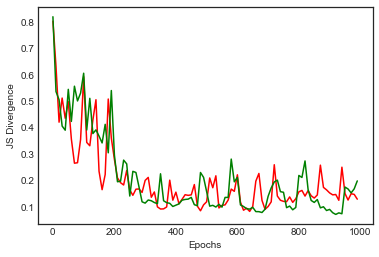}} &
\subfloat[Overlapping Ellipses]{\includegraphics[width = 1.5in]{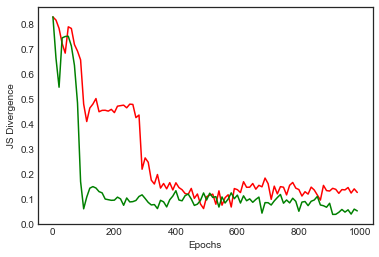}} &
\subfloat[Concentric Circles]{\includegraphics[width =1.5in]{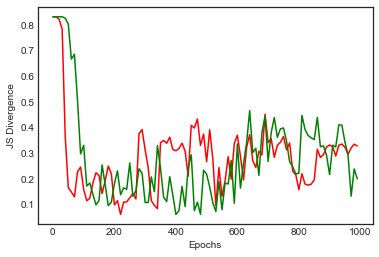}} 
\end{tabular}
\caption{Jensen–Shannon Divergence}
\label{jsdiverge}
\end{figure}

The Jensen–Shannon divergence is a  statistical method of measuring the similarity between two probability distributions. It is based on the Kullback–Leibler divergence.  The main difference is it is symmetric and it is always between zero and one. In Figure \ref{jsdiverge}, red coloured graphs show the JS divergence  between the sample generated by  GAN and   the ground-truth data set. The  green coloured graphs show the JS divergence between  the sample generated by CEN compared and the ground-truth data set. 

\begin{figure}
\begin{tabular}{ccc}
\subfloat[Sine Wave]{\includegraphics[width = 1.5in]{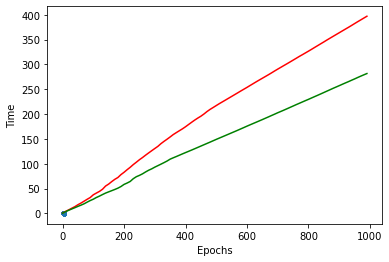}} &
\subfloat[Overlapping Ellipses]{\includegraphics[width = 1.5in]{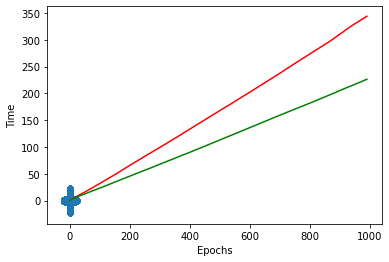}} &
\subfloat[Concentric Circles]{\includegraphics[width =1.5in]{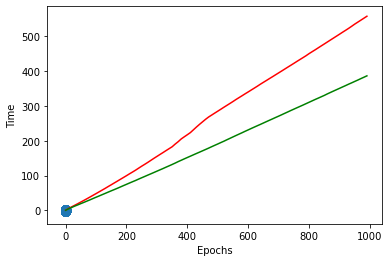}} 
\end{tabular}
\caption{Computation Time}
\label{comptime}
\end{figure}

IN Figure \ref{comptime}, red coloured graphs show the required computation time for  GAN. The  green coloured graphs show the required computation time for CEN. We can see CEN achieved a significant improvement. 

\subsection{MNIST handwritten digits}
\begin{figure}
\begin{tabular}{cccc}
\subfloat[100 Epochs]{\includegraphics[width = 0.9in]{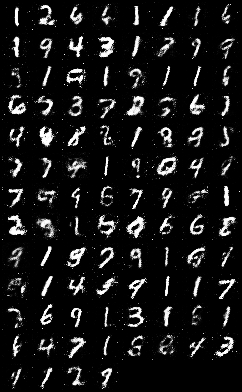}} &
\subfloat[200 Epochs]{\includegraphics[width = 0.9in]{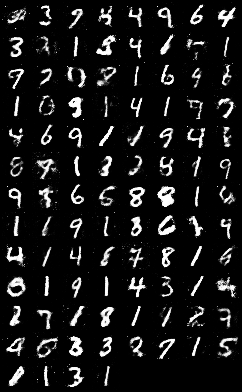}} &
\subfloat[300 Epochs]{\includegraphics[width = 0.9in]{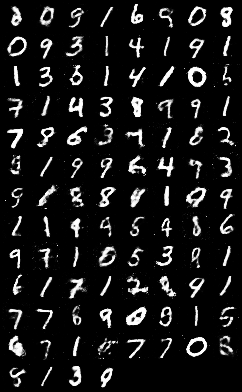}} &
\subfloat[400 Epochs]{\includegraphics[width = 0.9in]{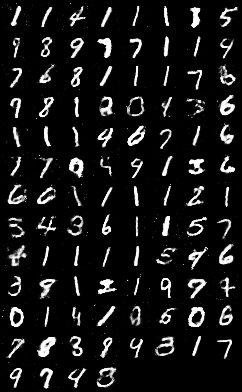}}\\
\subfloat[500 Epochs]{\includegraphics[width = 0.9in]{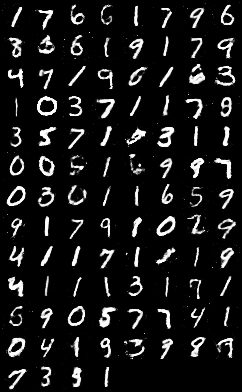}} &
\subfloat[600 Epochs]{\includegraphics[width = 0.9in]{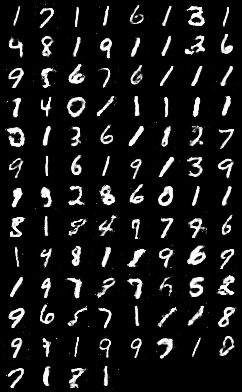}} &
\subfloat[700 Epochs]{\includegraphics[width = 0.9in]{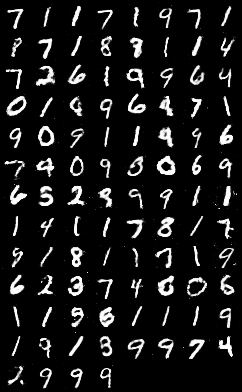}} &
\subfloat[800 Epochs]{\includegraphics[width = 0.9in]{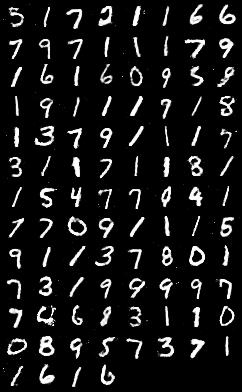}}
\end{tabular}
\caption{Generated by GAN}
\label{G4GAN}
\end{figure}

\begin{figure}
\begin{tabular}{cccc}
\subfloat[100 Epochs]{\includegraphics[width = 0.9in]{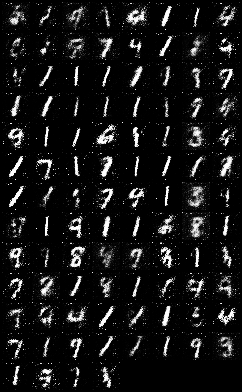}} &
\subfloat[200 Epochs]{\includegraphics[width = 0.9in]{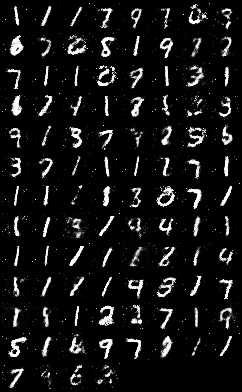}} &
\subfloat[300 Epochs]{\includegraphics[width = 0.9in]{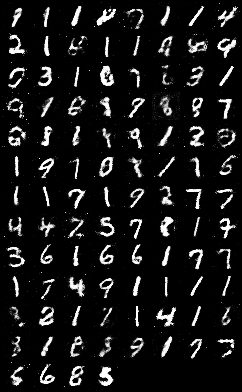}} &
\subfloat[400 Epochs]{\includegraphics[width = 0.9in]{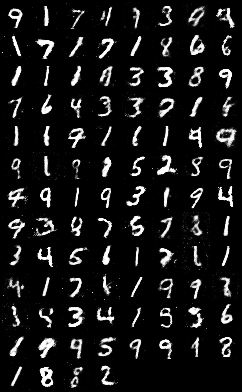}}\\
\subfloat[500 Epochs]{\includegraphics[width = 0.9in]{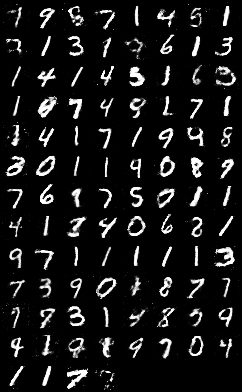}} &
\subfloat[600 Epochs]{\includegraphics[width = 0.9in]{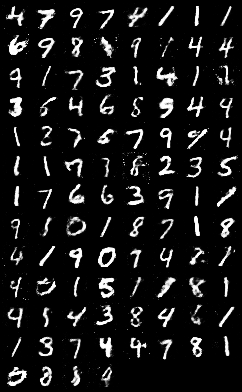}} &
\subfloat[700 Epochs]{\includegraphics[width = 0.9in]{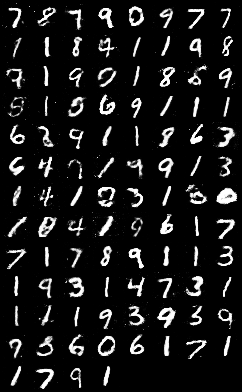}} &
\subfloat[800 Epochs]{\includegraphics[width = 0.9in]{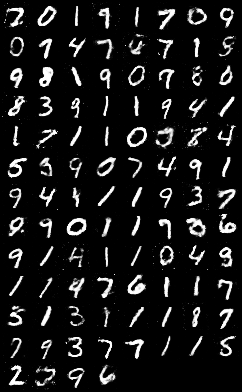}}

\end{tabular}
\caption{Generated by CEN}
\label{C4CEN}
\end{figure}

The ground-truth data set is the MNIST database of handwritten digits.  The Figure \ref{G4GAN} shows the sample data generated for different numbers of epchos (from one hundred to eight hundred) by GAN. The Figure \ref{C4CEN} shows the sample data generated for different number of epchos (from one hundred to eight hundred) by CEN. 

\subsection{MNIST handwritten digits one, two and three}
\begin{figure}
\begin{tabular}{cccc}
\subfloat[100 Epochs]{\includegraphics[width = 0.9in]{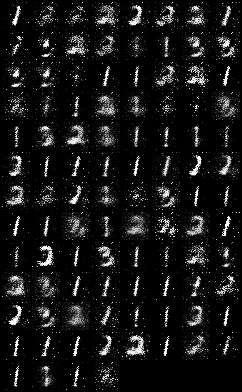}} &
\subfloat[200 Epochs]{\includegraphics[width = 0.9in]{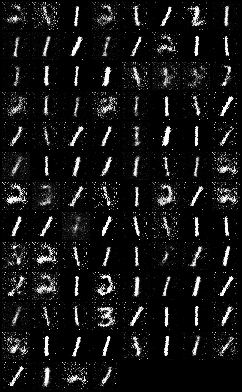}} &
\subfloat[300 Epochs]{\includegraphics[width = 0.9in]{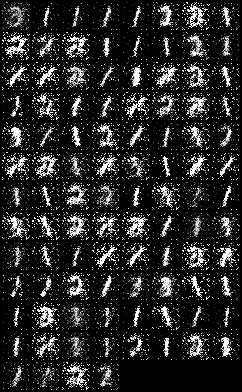}} &
\subfloat[400 Epochs]{\includegraphics[width = 0.9in]{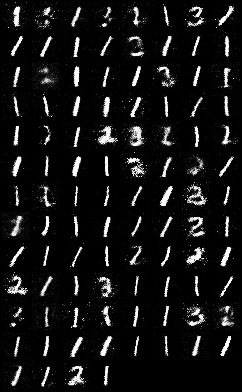}}\\
\subfloat[500 Epochs]{\includegraphics[width = 0.9in]{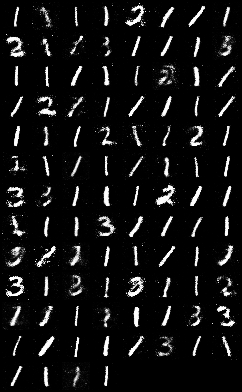}} &
\subfloat[600 Epochs]{\includegraphics[width = 0.9in]{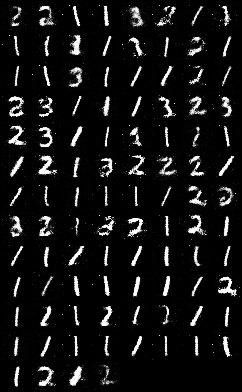}} &
\subfloat[700 Epochs]{\includegraphics[width = 0.9in]{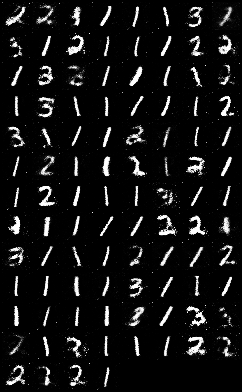}} &
\subfloat[800 Epochs]{\includegraphics[width = 0.9in]{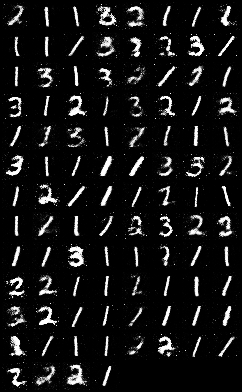}}
\end{tabular}
\caption{Generated by GAN}
\label{G5GAN}
\end{figure}

\begin{figure}
\begin{tabular}{cccc}
\subfloat[100 Epochs]{\includegraphics[width = 0.9in]{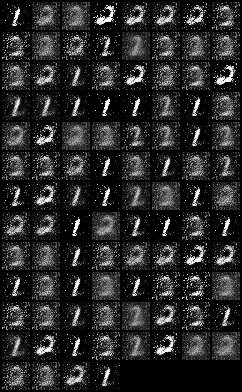}} &
\subfloat[200 Epochs]{\includegraphics[width = 0.9in]{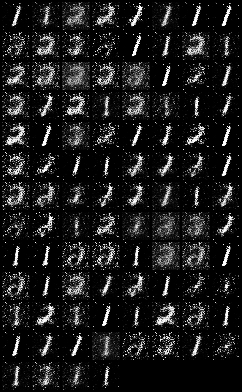}} &
\subfloat[300 Epochs]{\includegraphics[width = 0.9in]{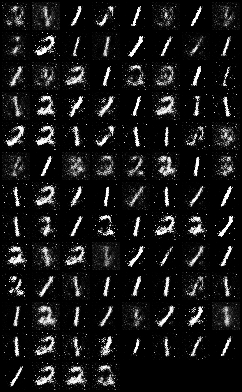}} &
\subfloat[400 Epochs]{\includegraphics[width = 0.9in]{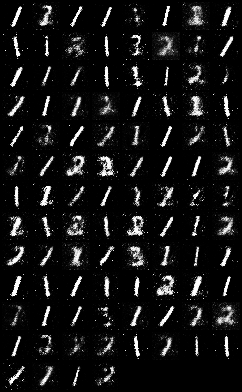}}\\
\subfloat[500 Epochs]{\includegraphics[width = 0.9in]{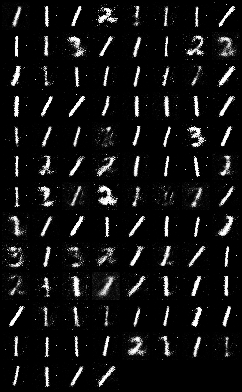}} &
\subfloat[600 Epochs]{\includegraphics[width = 0.9in]{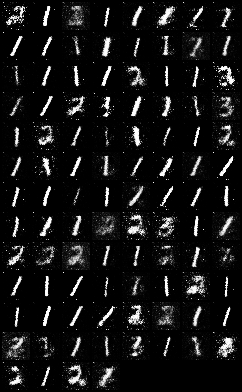}} &
\subfloat[700 Epochs]{\includegraphics[width = 0.9in]{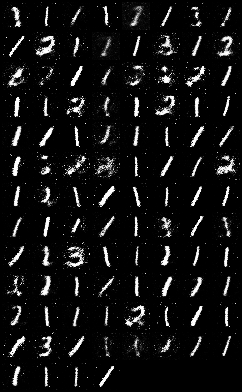}} &
\subfloat[800 Epochs]{\includegraphics[width = 0.9in]{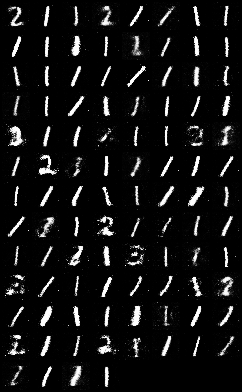}}

\end{tabular}
\caption{Generated by CEN}
\label{C5CEN}
\end{figure}

The ground-truth data set is the MNIST database of handwritten digits. We had taken five thousand images of digit one, three thousand images of digit two and two thousand images of digit three. The Figure \ref{G5GAN} shows the sample data generated for different numbers of epchos (from one hundred to eight hundred) by GAN. The Figure \ref{C5CEN} shows the sample data generated for different number of epchos (from one hundred to eight hundred) by CEN. 

\section{Conclusion}
The training procedure of GAN is modeled as a finitely repeated simultaneous game. Each module tries to increase its performance at every repetition of the base game in a non-cooperative manner. We showed experimentally and statistically  that each module can perform better and learn faster (around 25\% reduction in the training time) if training is modeled as an infinitely repeated simultaneous game. At every repetition of the base game, the stronger module cooperates with the weaker module and only the weaker module is allowed to increase its performance. 

\bibliographystyle{splncs04}
\bibliography{main}
\end{document}